\documentclass{article}

\usepackage{booktabs}
\usepackage{multirow}
\usepackage{multicol}
\usepackage{titletoc}
\usepackage{amsmath}  
\usepackage{bbding}
\usepackage{amssymb}  %
\usepackage[table]{xcolor}

\usepackage{microtype}
\usepackage{graphicx}
\usepackage{subcaption}
\usepackage{booktabs} 

\usepackage{hyperref}

\usepackage[preprint]{icml2026}

\usepackage{amsmath}
\usepackage{amssymb}
\usepackage{mathtools}
\usepackage{amsthm}

\usepackage[capitalize,noabbrev]{cleveref}

\theoremstyle{plain}

\theoremstyle{definition}

\theoremstyle{remark}

\usepackage[textsize=tiny]{todonotes}

\icmltitlerunning{VideoAfford: Grounding 3D Affordance from Human-Object-Interaction Videos via Multimodal Large Language Model}

\begin{document}

\twocolumn[
  \icmltitle{VideoAfford: Grounding 3D Affordance from Human-Object-Interaction Videos \\via Multimodal Large Language Model}





\newcommand{\affilhkustgz}{$^1$}
\newcommand{\affilzju}{$^2$}
\newcommand{\affilskd}{$^3$}
\newcommand{\affilhust}{$^4$}
\newcommand{\affilsdu}{$^5$}
\newcommand{\affilhkust}{$^6$}
\newcommand{\caffilhkust}{$^{,6}$}

\newcommand{\corrmark}{$^{,\dagger}$}      

\begin{center}
{\bf Hanqing Wang\affilhkustgz} \quad
{\bf Mingyu Liu\affilzju} \quad
{\bf Xiaoyu Chen\affilhkustgz} \quad
{\bf Chengwei Ma\affilhkustgz} \quad
{\bf Yiming Zhong\affilskd} \quad
{\bf Wenti Yin\affilhust} \quad
{\bf Yuhao Liu\affilsdu} \quad
{\bf Zhiqing Cui\affilhkustgz} \quad
{\bf Jiahao Yuan\affilhkustgz} \quad
{\bf Lu Dai\affilhkustgz\caffilhkust} \quad
{\bf Zhiyuan Ma\affilhust\corrmark} \quad
{\bf Hui Xiong\affilhkustgz\caffilhkust\corrmark} \quad

\vspace{0.15in}

{\small
$^1$HKUST-GZ\quad
$^2$ZJU\quad
$^3$ShanghaiTech \quad
$^4$HUST \quad
$^5$SDU \quad
$^6$HKUST \quad

}

\vspace{0.08in}

\end{center}

  \vskip 0.3in
  {
\renewcommand\twocolumn[1][]{#1}

\begin{center}
    \captionsetup{type=figure}
    \includegraphics[width=\linewidth]{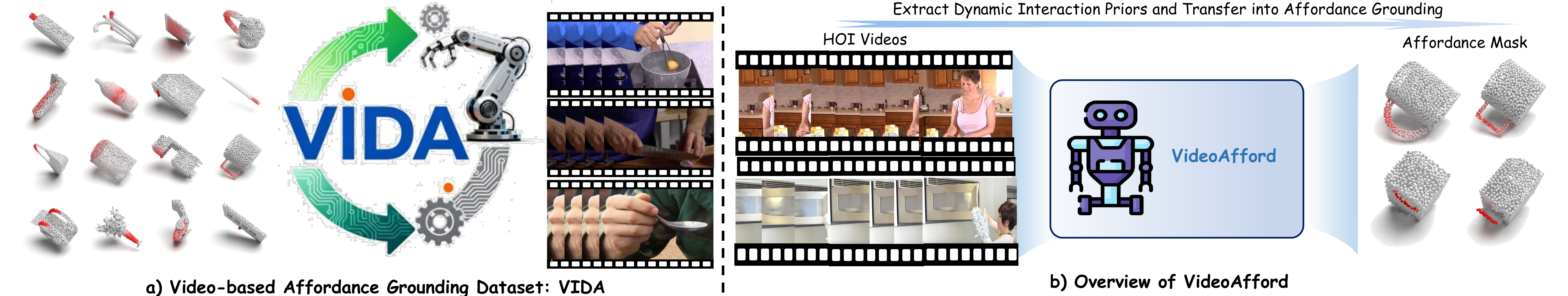}
     \caption{We introduce the task of grounding 3D affordance from human-object-interaction (HOI) videos, and collect a large-scale dataset VIDA (left), which contains 38K HOI videos spanning 16 affordance categories and 22K 3D point clouds. Based on this, we proposed the VideoAfford (right), which can transfer the HOI interaction priors into 3D affordance grounding.}
     \label{figvis}

\end{center}
}
]



\printAffiliationsAndNotice{}  

\begin{abstract}
3D affordance grounding aims to highlight the actionable regions on 3D objects, which is crucial for robotic manipulation. Previous research primarily focused on learning affordance knowledge from static cues such as language and images, which struggle to provide sufficient dynamic interaction context that can reveal temporal and causal cues. To alleviate this predicament, we collect a comprehensive video-based 3D affordance dataset, \textit{VIDA}, which contains 38K human-object-interaction videos covering 16 affordance types, 38 object categories, and 22K point clouds. Based on \textit{VIDA}, we propose a strong baseline: VideoAfford, which activates multimodal large language models with additional affordance segmentation capabilities, enabling both world knowledge reasoning and fine-grained affordance grounding within a unified framework. To enhance action understanding capability, we leverage a latent action encoder to extract dynamic interaction priors from HOI videos. Moreover, we introduce a \textit{spatial-aware} loss function to enable VideoAfford to obtain comprehensive 3D spatial knowledge.  Extensive experimental evaluations demonstrate that our model significantly outperforms well-established methods and exhibits strong open-world generalization with affordance reasoning abilities. All datasets and code will be publicly released to advance research in this area.
\end{abstract}

\section{Introduction}
\label{sec:intro}

Recent years have witnessed the remarkable success of affordance~\cite{gibson1977theory} grounding, particularly in a wide variety of embodied intelligence applications.
Affordance grounding aims to identify and locate the operable areas of an object to bridge visual perception and robotic manipulation, in response to human commands or demonstrations.
Previous studies have focused primarily on 2D affordance segmentation from visual modalities, including images~\cite{thermos2020deep,do2018affordancenet,luo2024visual,zhao2020object} and videos~\cite{bahl2023affordances,luo2023learning,chen2023affordance}. 
To enhance open-vocabulary capabilities, subsequent studies~\cite{nguyen2023open,tang2025uad,jiang2025affordancesam,chen2024worldafford，zhang2025a4} have leveraged foundation models such as CLIP~\cite{radford2021learning} to incorporate linguistic information into the affordance grounding pipeline. More recently, AffordanceLLM~\cite{qian2024affordancellm} has advanced this direction by harnessing the world knowledge encoded in multimodal large language models for affordance reasoning. 
Motivated by \emph{learning-from-demonstration} paradigms, another line of research~\cite{tang2025closed,luo2022learning,xuweakly,li2023locate,wang2025reasoning,moon2025selective} has exploited human-object-interaction (HOI) images as interaction priors to facilitate affordance localization. While 2D affordance representations provide valuable visual cues suggesting potential actions for embodied systems, 3D affordance offers a more precise and intuitive guidance for task execution in realistic physical environments, thereby establishing a robust foundation for a wide range of downstream embodied AI applications, including robotic grasping~\cite{wei2025afforddexgrasp,zhao2025towards,zhang2023affordance,9981900} and manipulation~\cite{wu2023learning, wu2023learning2,wu2025garmentpile,xu2024naturalvlm}.
Despite these advances, existing approaches still significantly rely on static information sources such as HOI images~\cite{GREAT,IAGNet,gao2024learning,yang2024lemon} or linguistic instructions~\cite{laso,SeqAfford,chu20253d,delitzas2024scenefun3d,3DAffordSplat,SeqAffordSplat} for 3D affordance learning. Such static modalities inherently lack the temporal dynamics required to capture interaction patterns — a critical element for understanding the causal mechanisms underlying affordance. Several recent efforts have explored learning from HOI videos~\cite{bahl2023affordances,luo2023learning,chen2023affordance,fang2018demo2vec,heidinger20252handedafforder,liu2024grounding} to alleviate this predicament. Specifically, OPPA~\cite{fang2018demo2vec} curates a diverse collection of internet videos and establishes a benchmark for 2D affordance perception. VRB~\cite{bahl2023affordances} exploits the rich representational capacity of HOI videos to enhance affordance-driven robotic manipulation. Similarly, 2HandedAfforder~\cite{heidinger20252handedafforder} develops an automated annotation system for video-based affordance labels and their subsequent transfer to dual-arm robotic systems. Nevertheless, these methods either demand labor-intensive manual annotations or remain confined to 2D representations. More recently, EGO-SAG~\cite{liu2024grounding} has pioneered the learning of 3D affordance from unlabeled egocentric video data. However, its focus on scene-level coarse-grained masks falls short of the object-centric fine-grained affordance segmentation required for precise robotic manipulation and grasping.

\begin{figure*}[t]
    \centering
    \includegraphics[width=0.9\linewidth]{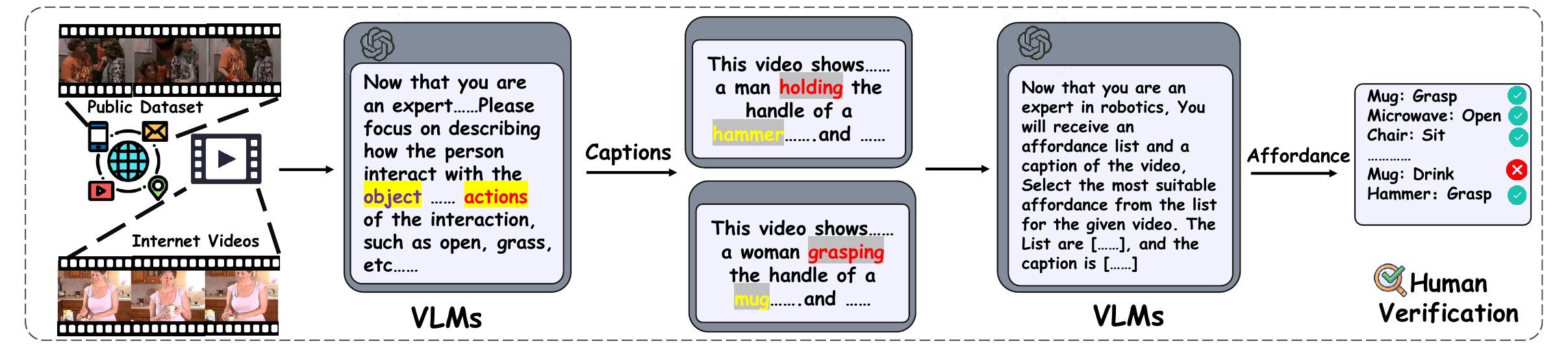}
    \caption{\textbf{Data Collection Pipeline.} We show the whole data collection and verification pipeline here. First, we utilize VLMs to caption each video and extract keywords about action and objects. We then utilize the VLMs to pair the video to an affordance type. Finally, we manually check the results to ensure correctness.}
    \label{fig:data_collect}
\end{figure*}

To address the above limitations, we formulate a novel task: \textit{Grounding 3D object affordance from human-object interaction demonstration videos}, which aims to harness large-scale demonstration video corpora for object-centric 3D affordance reasoning. To this end, we collect and construct \textbf{VIDA}, a large-scale video-point cloud pair dataset. Our data collection pipeline aggregates HOI videos from diverse sources, including internet repositories like YouTube, HOIGEN-1M~\cite{liu2025hoigen}, and TASTE-RoB~\cite{zhao2025taste}. As depicted in Fig.~\ref{fig:data_collect}, we employ state-of-the-art Vision-Language Models (VLMs), specifically GPT-4o~\cite{achiam2023gpt}, to generate descriptive captions for internet videos, from which we extract object and action information. Subsequently, VLMs analyze the correspondence between video actions and affordance categories. The resulting annotations undergo rigorous manual verification and correction to ensure quality. Motivated by the remarkable comprehension capabilities of Multimodal Large Language Models (MLLMs) in visual understanding, we observe that video MLLMs possess the inherent ability to recognize interactions and localize affordance regions within videos, reflecting their extensive world knowledge. This observation inspires us to unlock and transfer the affordance reasoning capabilities embedded in video MLLMs to the 3D affordance grounding domain. Building upon VIDA, we propose \textbf{VideoAfford}, a strong baseline that leverages the world knowledge inherent in video MLLMs. To endow our model with dynamic action comprehension, we incorporate a latent action encoder that extracts action embeddings from HOI videos. Furthermore, we introduce a spatial-aware loss mechanism to enhance the model's spatial reasoning capabilities, enabling \textbf{VideoAfford} to acquire comprehensive 3D spatial knowledge. The synergistic integration of these components enables \textbf{VideoAfford} to achieve superior performance on both in-distribution and out-of-distribution data, a critical requirement for practical deployment in embodied perception systems. Our contributions can be summarized:
\begin{itemize}
    \item We formulate the task of video-based 3D object affordance grounding, which aims to unleash the potential within human-object-interaction videos and transfer rich interaction priors for 3D affordance grounding. 

    \item To support this task, we present \textit{\textbf{VIDA}}, the first large-scale video-based affordance grounding benchmark  comprising 38K HOI videos spanning 16 affordance categories and 22K annotated affordance point clouds. \textit{\textbf{VIDA}} will be made publicly available to facilitate future research in this domain.
    
    \item We propose \textit{\textbf{VideoAfford}}, a robust baseline designed to effectively exploit the affordance knowledge embedded in video data through the seamless integration of spatial-aware loss functions and latent action encoding mechanisms.
    
    \item Extensive experimental results demonstrate substantial performance gains of our method over existing baselines while exhibiting robust open-vocabulary generalization, thereby confirming the practical viability of our method for real-world deployment.
\end{itemize}
\section{Related Works}
\label{sec:related_works}

\paragraph{Affordance Learning.} Affordance is a critical point for linking perception, reasoning, and manipulation in the real physical world. Some works are devoted to detecting the affordance region from 2D sources, i.e., images~\cite{thermos2020deep,do2018affordancenet,luo2024visual,zhao2020object, wang2025affordance} and videos~\cite{bahl2023affordances,luo2023learning,chen2023affordance}, and there are also some studies that accomplish this task with the assistance of natural language~\cite{chen2024worldafford}. They seek to bring a leap to locate the affordance region of objects from 2D sources. However, the affordance knowledge derived from 2D sources is difficult to transfer to specific interactive locations of objects in the 3D environment. To address this gap, 3D AffordanceNet~\cite{deng20213d} first proposed a large-scale fine-grained 3D affordance dataset, which has greatly promoted the advancement in 3D affordance learning. Based on this, many researchers have further explored methods for learning universal affordance knowledge from HOI images~\cite{GREAT,IAGNet,gao2024learning,yang2024lemon} and language instructions~\cite{tian20253,laso, SeqAfford,chu20253d,delitzas2024scenefun3d,3DAffordSplat, SeqAffordSplat, wang2025affordbot, liu2025pavlm}. Some research has also proposed utilizing the rich world knowledge within foundation models, such as vision language models~\cite{SeqAfford,chu20253d} or diffusion models~\cite{wang2025dag,ju2024robo}, to assist 3D affordance learning. However, these works overlook the dynamic interaction of information, which is crucial for affordance reasoning. To overcome this, EGO-SAG~\cite{liu2024grounding} proposed learning interaction affordance knowledge from egocentric views, but it mainly focuses on scene affordance and does not provide rich affordance prior knowledge for object-centric manipulation. To fill this gap, we collect a large number of HOI videos and paired 3D point clouds to construct the first video-based object-centric 3D affordance learning dataset \textbf{VIDA}. VIDA aims to advance 3D affordance research to better promote embodied intelligence.

\paragraph{Multimodal Large Language Model.}
Multimodal Large Language Models (MLLMs) have demonstrated impressive abilities in 2D~\cite{achiam2023gpt,yang2025qwen3,zhu2023minigpt} and 3D understanding~\cite{xu2024pointllm,qi2024shapellm,hong20233d}. Trained on large-scale web data, MLLMs exhibit human-like reasoning and perception across diverse modalities.
However, their potential for embodied perception—especially in affordance reasoning—remains largely untapped. Recent efforts such as AffordanceLLM~\cite{qian2024affordancellm} and 3DAffordanceLLM~\cite{chu20253d} have begun to explore affordance grounding utilizing MLLMs, but most rely on static images or text and overlook dynamic interaction cues crucial for real-world manipulation. To address this issue, our work aims to endow MLLMs with affordance-aware perception and reasoning abilities by learning from video-based human–object interactions, enabling them to interpret and act upon 3D objects in context-sensitive scenarios.

\paragraph{3D Spatial Reasoning.}Recently, 3D spatial reasoning has attracted extensive attention from both academic and industrial communities, serving as a critical driver for advancing the capabilities of Embodied AI in perceiving and understanding the physical world. Existing studies have explored multiple avenues to enhance models' comprehension of 3D objects: one line of research~\cite{zhang2023uni3d,wang2025partnext,zhou2024point,yang2024sampart3d,yang2023sam3d} improves the understanding of 3D objects within specific semantic categories by segmenting their semantic parts, yet they struggle to generalize to unseen categories; another line of research~\cite{hong20233d,qu2025spatialvla,ma2025spatialllm,wang2025odyssey,liu2025bridge} leverages large language models (LLMs) to map rich semantic features to 3D objects, thereby enabling agents to grasp object functions and geometric structures. However, a notable limitation of these methods lies in their primary focus on static mapping—they largely neglect the interactivity inherent in real-world environments, failing to capture the dynamic interaction affordances of 3D objects. To address this gap between static perception and dynamic interaction, this paper proposes a novel approach that grounds object affordances in 3D spaces through HOI videos.


\section{Datasets}

\label{sec:dataset}
\begin{table*}[t]
\centering
\caption{\footnotesize
\textbf{Comparison of Existing 3D Affordance Datasets with Ours.}
\#Point Cloud, \#Aff, and \#Classes denote the number of point clouds, affordance types, and objects, respectively. \#Dynamic Information means if the dataset can provide the dynamic interaction information to assist affordance grounding. $\checkmark$/$\times$ indicates that if the dataset can possess this attribute.}
\small 
\resizebox{\linewidth}{!}{
\begin{tabular}{lccccccc}
\toprule
\textbf{Dataset} & \textbf{\#Publication Year} &\textbf{\#Dynamic Information} & \textbf{\#Granularity} &  \textbf{\#Input Source} &  \textbf{\#Classes} & \textbf{\#Aff} & \textbf{\#Point Cloud} \\
\midrule
3D AffordanceNet~\cite{Deng_2021_CVPR} &CVPR 2021 & $\times$ &Object & Point Cloud & 23& 18 & 23k \\
O2O-Afford~\cite{Mo_2021_CoRL} &CoRL  2021 & $\times$  &Object & Point Cloud & 18 & -- & 1.7k \\
PartAfford~\cite{Xu_2022_arXiv} &ECCVW 2022 & $\times$  &Object & Point Cloud & 23 & 24 & 25k \\
IAGNet~\cite{IAGNet} &ICCV 2023 & $\times$  &Object & Image, Point Cloud & 23 & 17 & 7k \\
LASO~\cite{laso} &CVPR 2025 & $\times$  &Object & Language, Point Cloud & 23 & 17 & 8.4k \\

SceneFun3D~\cite{delitzas2024scenefun3d} &CVPR 2024 & $\times$  &Scene & Language, Point Cloud  & -- & 9 & 710  \\

SeqAfford~\cite{SeqAfford} &CVPR 2025 & $\times$  &Object & Language, Point Cloud  & 23 & 28 & 180k \\

3DAffordSplat~\cite{3DAffordSplat} &ACM MM 2025 & $\times$  &Object & Language, 3D Gaussian  & 21 & 18 & 24k \\
SeqAffordSplat~\cite{SeqAffordSplat} &Arxiv 2025 & $\times$  &Object & Language, 3D Gaussian & 21 & 18 & 14k \\
GREAT~\cite{GREAT} &CVPR 2025 & $\times$  &Object & Language,Image, Point Cloud  & 43 & 24 & 39k \\

AGPIL~\cite{AGPIL} &CVPR 2025 & $\times$  &Object &  Language,Image, Point Cloud & 23 & 17 & 31k \\

Affogato~\cite{Affogato} &Arxiv 2025  & $\times$  &Object &  Language,Image, Point Cloud & \textgreater 450 & \textgreater 350 & 150k \\
\midrule
VSAD~\cite{liu2024grounding} &Arxiv 2024& \checkmark  &Scene  & Videos, Point Cloud & 16 & 17 & 2k \\
\textbf{VIDA} (Ours) &2025& \checkmark  &Object  & Language, Videos, Point Cloud & 38 & 16 & 22k \\
\bottomrule
\end{tabular}
}

\label{tab:dataset_comparison}
\end{table*}

\begin{figure*}[t]
    \centering
    \includegraphics[width=0.95\linewidth]{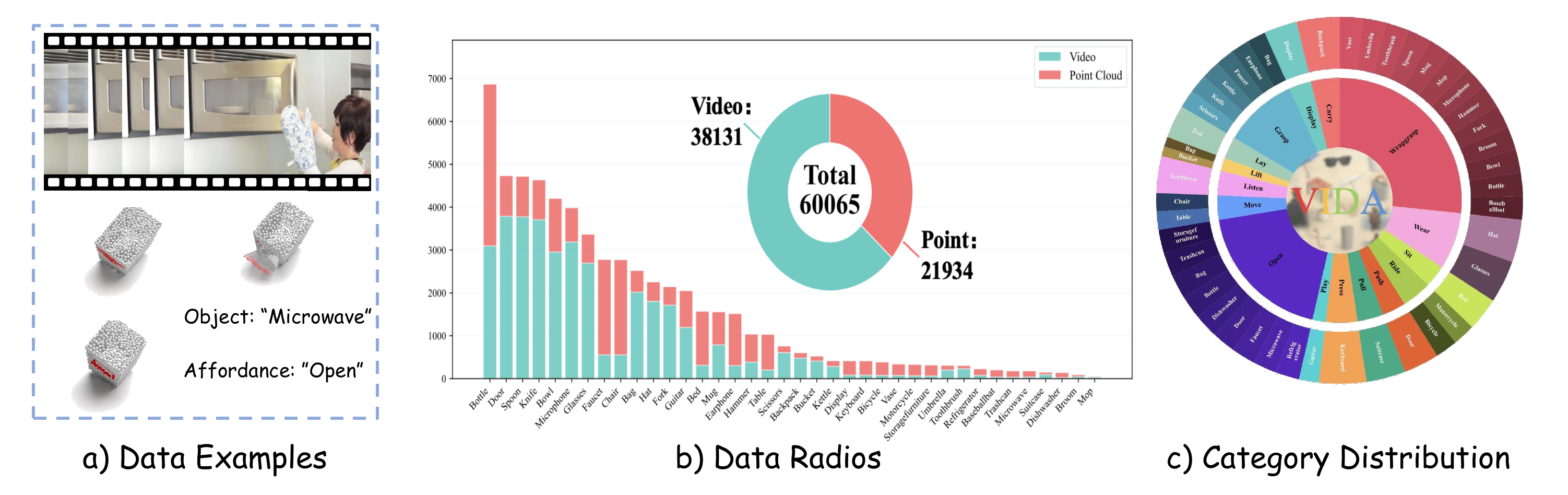}
    \caption{\textbf{VIDA Dataset.} Here we illustrate the detailed information of VIDA. a) shows the examples of the video and corresponding affordance point clouds. b) shows the videos and point clouds radios, and c) shows the category distributions of VIDA.}
    \label{fig:data_all}
\end{figure*}

Previous affordance datasets mainly focus on static images or texts, and fail to provide complex dynamic interaction priors. The pre-contact, contact, and post-contact frames in HOI videos can provide rich temporal information about the intention of the interaction and the causal relationship of the contact. To the best of our knowledge, \textit{\textbf{VIDA}} is the first large benchmark designed to unleash the potential of HOI videos for the 3D object-centric affordance grounding. 

\subsection{Collection Details}
\paragraph{Videos and Point Clouds.} 
The \textit{\textbf{VIDA}} dataset contains approximately \textbf{38K} HOI videos collected from public data sources like HOIGEN-1M~\cite{liu2025hoigen}, TASTE-Rob~\cite{zhao2025taste}, and the Internet resources, covering \textbf{38} object categories and 16 affordance types (e.g., \emph{open}, \emph{push}). We first filter the captions of each video by extracting the corresponding objects and actions. The GPT-4o~\cite{achiam2023gpt} is exploited to further analyze which affordance corresponds to the current action for each video. Finally, we manually check the results to ensure the correctness of each video. Moreover, we further collected the point clouds from PIADv1~\cite{yang2023grounding3dobjectaffordance} and PIADv2~\cite{GREAT}, then we solely selected those point cloud objects that can be paired with the HOI videos for dataset construction and experiments.

\subsection{Statistic and Analysis}
As illustrated in Fig. \ref{fig:data_all}, \textit{\textbf{VIDA}} dataset covers over $38.1$k videos and $21.9$k object point clouds, which encompasses 38 object categories and 16 affordance types. Since videos and point clouds are sampled from different instances, for training, they do not need a fixed one-to-one pairing; one video could be paired with multiple point clouds, and the number of them is not strictly consistent. An example can be seen in Fig. \ref{fig:data_all} (a), a video can correspond to multiple point clouds. While for evaluation, we strictly select videos and point clouds one-on-one to ensure the rigor and reproducibility of the evaluation results.  Fig. \ref{fig:data_all} (b) illustrates the ratio of videos and point clouds in each affordance category.  Fig. \ref{fig:data_all} (c) shows the count and distribution of affordances in videos and point clouds. Following PIADv1~\cite{yang2023grounding3dobjectaffordance} and PIADv2~\cite{GREAT}, we divide the VIDA into seen and unseen settings. In the seen setting, both objects and affordances in the training and testing sets are consistent, while in the unseen setting, some objects or affordances in the testing set do not exist in the training set.

\begin{figure*}[t]
    \centering
    \includegraphics[width=0.95\linewidth]{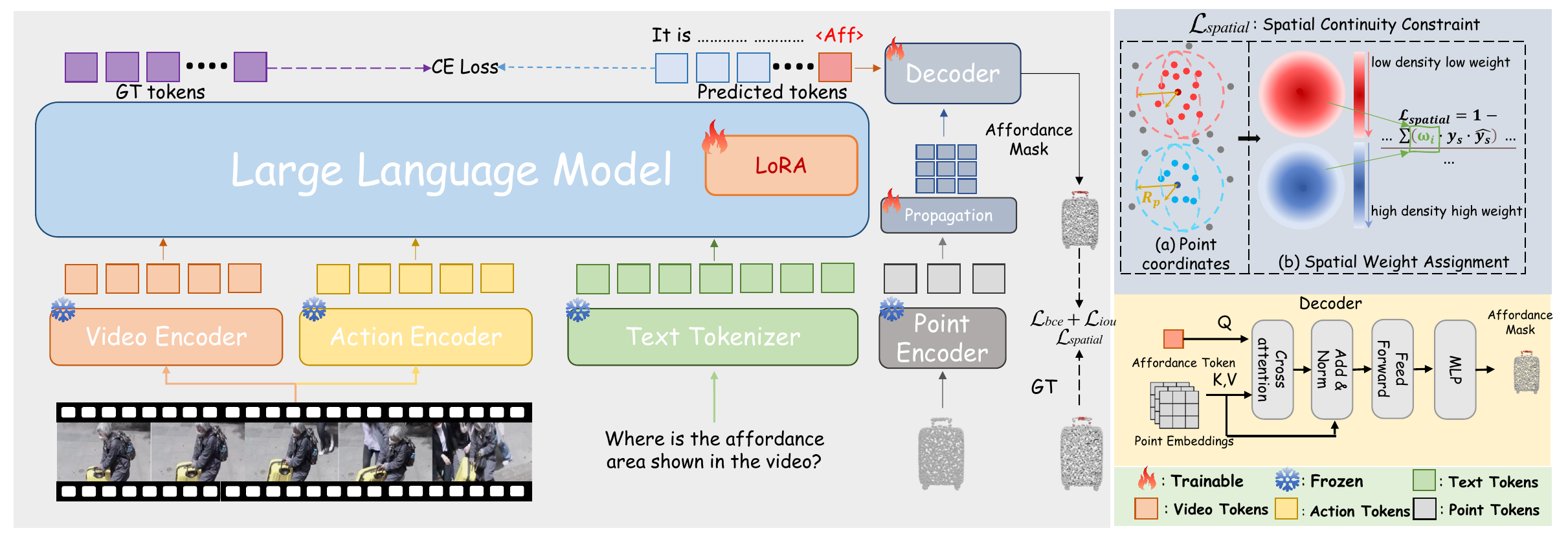}
    \caption{\textbf{Overview of VideoAfford.} Given an HOI video and a corresponding point cloud, VideoAfford adopts the LanguageBind as the video encoder and RenderNet as the action encoder to obtain the video embeddings and latent action embeddings. Then the video embeddings and latent action embeddings are fed into the Large Language Model to predict the language tokens and the affordance token. On the other hand, VideoAfford utilizes a pre-trained 3D encoder to extract the semantic-rich point embeddings, which are then fed into a geometric guided upsample and propagation module to obtain dense point features. Finally, the affordance token and the point features are fed into the affordance decoder to obtain the affordance masks. More details about the propagation process can be seen in appendix.}
    \label{fig:pipeline}
\end{figure*}
\section{Methods}
In this section, we give the details of our proposed framework: VideoAfford.  We first give an overview in Sec.~\ref{overview}, giving the definition of the proposed task. We then detail our network architecture in Sec.~\ref{Network Architecture}. Finally, we provide the training objectives in Sec.~\ref{training objectives}.

\subsection{Architecture Overview}
\label{overview}
We provide an overview of our proposed method \textbf{VideoAfford}. We reformulate the affordance grounding task as learning affordances from HOI video demonstrations, where the model aims to localize actionable areas on 3D objects based on the given HOI videos. Formally, given an HOI video $\mathcal{V}$ and a text instruction $\mathcal{T}$, the model $g$ is expected to output the affordance mask $\mathcal{A}_f$, defined as:

\begin{equation}
\mathcal{A}_f=g(\mathcal{T},\mathcal{V}).
\end{equation}
As shown in Fig.\ref{fig:pipeline}, \textbf{VideoAfford} mainly consists of four components: 1) a 3D vision encoder benefited from large-scale 3D representation learning, which provides solid foundations for dense prediction tasks; 2) a pre-trained latent action encoder, which provides rich action priors; 3) a video multimodal large language model $g$ that exhibits affordance reasoning ability with the aid of internalized world knowledge; 4) A transformer-based lightweight affordance decoder that integrates affordance embeddings into point embeddings to predict affordance masks, thereby enabling fine-grained and spatially consistent affordance understanding.

\subsection{Network Architecture}
\label{Network Architecture}

 \paragraph{Point Encoder.} Inspired by the previous works, we adopt a pre-trained 3D encoder, which has been trained on large \emph{text-image-point} paired data, as our 3D point cloud backbone. Given the point cloud $\mathcal{P}_c \in \mathcal{R}^{N \times 3}$, the 3D point cloud encoder $f$ encodes the point cloud into semantic-rich point embeddings $\mathcal{P}_s$. Then, following Point-Bert~\cite{yu2022point}, we perform geometric guided up-sampling to propagate the semantics features into dense point features $\mathcal{P}_d \in \mathcal{R}^{N \times 2048\times 3}$, which can be formulated as:
  \begin{equation}
      \mathcal{P}_s  = f(\mathcal{P}_c),
 \end{equation}
   \begin{equation}
      \mathcal{P}_d  = Up(\mathcal{P}_s,o(\mathcal{P}_c)),
 \end{equation}
where $o$ represents geometric guided propagation. More details about the point encoder and upsample can be seen in our supplementary materials.

\paragraph{Spatial Constraints.} Most of previous works only focus on: \emph{``whether the predicted category for each point is accurate,'' } whereas lack constraints on \emph{``spatial continuity and regional overlapping.''} To address this limitation, we introduce the spatial loss, whose core mechanism incorporates spatial neighborhood information of point clouds to assign adaptive weights, thereby supporting the training of the spatial continuity. Specifically, we modified the Dice loss, utilizing point coordinates $\mathbf{x}_i \in \mathbb{R}^3$ (with all point coordinates denoted as $\mathcal{X} = \{\mathbf{x}_1, \mathbf{x}_2, ..., \mathbf{x}_N\} \in \mathbb{R}^{N \times 3}$) and a predefined radius $\mathcal{R}_p$ to identify neighboring points within the specified spatial range. Then we assign higher weights to spatially adjacent points, emphasizing the contribution of clustered positive samples during loss calculation:

\begin{equation}
\omega_i = \frac{1}{|\mathcal{N}_i|} \sum_{j \in \mathcal{N}_i} \exp\left(-\frac{\|\mathbf{x}_i - \mathbf{x}_j\|_2^2}{2\sigma^2}\right),
\end{equation}
where:
$\mathcal{N}_i = \left\{ j \mid \|\mathbf{x}_i - \mathbf{x}_j\|_2 \leq \mathcal{R}_p, j \neq i \right\}$ represents the spatial neighborhood of the $i$-th point, consisting of all points within a Euclidean distance $\mathcal{R}_p$ from $\mathbf{x}_i$;
$|\mathcal{N}_i|$ denotes the number of points in the neighborhood $\mathcal{N}_i$;
$\sigma$ is a hyper-parameter controlling the decay rate of weight with spatial distance (typically set as $\sigma = 0.1\mathcal{R}_p$);
$\omega_i \in \mathbb{R}^+$ is the adaptive spatial weight for the $i$-th point. By integrating the spatial weight $\omega_i$ into the traditional Dice loss, where $y_i \in [0,1]$ is the ground-truth label and $\hat{y}_i \in [0,1]$ is the predicted probability for the $i$-th point, we can derive:
\begin{equation}
\mathcal{L}_{spatial} = 1 - \frac{2 \sum_{i=1}^N \left( \omega_i \cdot y_i \cdot \hat{y}_i \right)}{\sum_{i=1}^N \left( \omega_i \cdot y_i^2 \right) + \sum_{i=1}^N \left( \omega_i \cdot \hat{y}_i^2 \right) + \epsilon}
\end{equation}
where $\epsilon$ is a small constant to avoid division by zero.
By introducing the aforementioned spatial loss, we compel the model to learn spatially continuous target regions, thus preventing fragmented predictions of intra-object points, which ensures that segmented targets maintain spatial compactness, aligning with the morphological characteristics of real-world objects.

\paragraph{Action Encoder.} To enhance the action understanding capability, inspired by \textbf{Richard Feynman}'s saying: \textit{What we observe as static is merely dynamic equilibrium}, we introduce the latent action encoder $m$ to learn generalizable human object interaction motions from a compact state representation. Specifically, given an HOI video, we sample N frames and use the action encoder to extract latent action embeddings and compress them into two tokens $\mathcal{A}_c \in \mathcal{R}^{N \times2 \times1024 }$, which can be formulated as follows:
     \begin{equation}
      \mathcal{A}_c  = m (\mathcal{V}).
 \end{equation}
 
\paragraph{Video MLLM Backbone.} We choose Video-LLaVA~\cite{lin2023video} as our
 backbone video multimodal large language model. Briefly, Video-LLaVA has both an image encoder and a video encoder, a text tokenizer, and a large language model LLM. The image encoder and the video encoder align images and videos before projection, allowing LLM to learn from a unified visual representation and endowing LLM with the ability to comprehend both images and videos simultaneously. When inputting a video $\mathcal{V}$ and a text $\mathcal{T}$, we use the video encoder $e$ and action encoder $m$ to encode the video sparsely to obtain video tokens and action tokens, and then concatenate them as the input for LLM. Then \textbf{VideoAfford} $g$ output the understanding text $\mathcal{J}$ as:
 \begin{equation}
      \mathcal{J}= g(m(V),e(\mathcal{V}), \mathcal{T}).
 \end{equation}
Following LISA~\cite{lai2024lisa}, we expand the vocabulary table of Video-LLaVA to inject the special token of $<AFF> $ to represent the affordance world knowledge, the hidden state of which is first projected into a query embedding $\mathcal{A}_m$ and then fed into a lightweight decoder as the affordance condition to generate a dense 3D affordance mask. 
\begin{equation}
    \mathcal{A}_m = proj(\mathcal{H}_\text{aff}).
\end{equation}

 \paragraph{Affordance Decoder.} To get dense affordance predictions, we propose a transformer-based light-weight decoder, which utilizes the affordance embedding $\mathcal{A}_m$ and the point features $\mathcal{P}_d $ to obtain the affordance mask $\mathcal{A}_{mask}$. We first fuse them by a cross-attention module to get $A_f$:
\begin{equation}
   \mathcal{A}_f =  \text{softmax}\left(\frac{\mathbf{Q} \cdot \mathbf{K}^T}{\sqrt{d}}\right) \cdot \mathbf{V}  ,
\end{equation}
where Q represents the affordance embedding $\mathcal{A}_m$,  K, V represent the point features $\mathcal{P}_d $. We finally get the affordance mask $\mathcal{A}_{mask}$ by inputting $\mathcal{A}_{f}$ into an MLP network.
\begin{equation}
  \mathcal{A}_{mask} = mlp(\mathcal{A}_{f}).
\end{equation}

\subsection{Training objectives.}
\label{training objectives}
Our strategy seeks to extract the rich affordance knowledge within HOI videos and transfer the rich priors into 3D affordance grounding in an end-to-end framework. Thus, we employ binary cross-entropy (BCE) and IOU loss to guide the segmentation mask prediction, and we also introduce a spatial loss to enhance the spatial understanding capability and ensure segmented targets maintain spatial compactness, consistent with the morphological characteristics of real-world objects. For the text output of language models, we follow the standard cross-entropy loss. To summarize, our final loss function is as follows:
\begin{equation}
\begin{aligned} 
 \mathcal{L} = \lambda_{ce}  \mathcal{L}_{ce} + \lambda_{bce}  \mathcal{L}_{bce} +  \lambda_{spatial}  \mathcal{L}_{spatial} \\ +  \lambda_{iou}  \mathcal{L}_{iou}
 \end{aligned} 
\end{equation}
where the weights $\lambda_{ce}, \lambda_{bce}, \lambda_{spatial}, \lambda_{iou}$ are utilized to balance the different loss items. 
\section{Experiments}

\begin{table}[t]
\centering
\caption{\textbf{Main Results.} The overall results of all comparative methods. AUC and mIoU are shown in percentage. The best results are in \textbf{bold} and the second results are in \underline{underline}.* means that we reproduce the code of the method.}
\begin{tabular}{@{}lcccccr@{}}
\toprule
& Method & \textit{mIoU}$\uparrow$ & \textit{AUC}$\uparrow$ & \textit{SIM}$\uparrow$ & \textit{MAE}$\downarrow$ \\
\midrule
\multirow{10}{*}{\rotatebox{90}{Seen}} 
& \multicolumn{5}{l}{\cellcolor[HTML]{F2F2F2}\textit{Non-MLLM-based Methods}} \\ 
& XMF   & 14.41 & 71.47 & 41.10 & 0.281 \\
& PFusion  & 16.33 & 78.43 & 46.28  & 0.264 \\
& IAGNet  & 20.39 & 80.22 & 50.11 & 0.188 \\
& LASO & 18.65 & 78.44 & 49.46 & 0.257 \\

& \multicolumn{5}{l}{\cellcolor[HTML]{F2F2F2}\textit{MLLM-based Methods}} \\ 
& GREAT  & \underline{23.62}   & \underline{81.41} & \underline{51.25} & \underline{0.173} \\
& Seqafford & 23.03 & 81.17 & 47.71 & 0.227 \\

& LMAfford3D*  & 22.74& 80.74 & 47.28 & 0.234 \\
& Ours & \textbf{28.20} & \textbf{83.64} & \textbf{58.80} & \textbf{0.157} \\
\midrule

\multirow{10}{*}{\rotatebox{90}{Unseen}} 
& \multicolumn{5}{l}{\cellcolor[HTML]{F2F2F2}\textit{Non-MLLM-based Methods}} \\

& XMF  & 6.010 & 53.41 &31.53 & 0.388 \\
& PFusion& 7.270 & 56.69 & 34.05 & 0.371 \\
& IAGNet   & 7.970 & 68.97 & 34.85 & 0.277 \\
& LASO  & 7.410 & 69.21 & 33.77 & 0.288 \\

& \multicolumn{5}{l}{\cellcolor[HTML]{F2F2F2}\textit{MLLM-based Methods}} \\ 
& GREAT  & \underline{8.220}   & \underline{70.19}  & \underline{35.08} & \underline{0.269} \\

& Seqafford  & 8.070 & 65.53 & 32.40 & 0.286 \\
& LMAfford3D*  &8.110 & 66.42 & 33.61 & 0.278\\

& Ours & \textbf{10.95} & \textbf{72.86} & \textbf{40.08} & \textbf{0.255} \\
\midrule
\bottomrule
\end{tabular}

\label{main_results}
\end{table}

\begin{table}[t]
\centering
\caption{We investigate the improvement of the Action Encoder and the proposed Spatial Constraint Loss on the model performance based on the baseline. }
\resizebox{\linewidth}{!}{
\begin{tabular}{c|cc|cccc}
\toprule
\multicolumn{1}{l|}{} & \textbf{Action Encoder} & \textbf{Spatial Loss}  & \textit{mIOU}$\uparrow$ &\textit{AUC}$\uparrow$ & \textit{SIM}$\uparrow$ & \textit{MAE}$\downarrow$ \\ 
\midrule
\multirow{4}{*}{\rotatebox{90}{Seen}} 
& $\times$ & $\times$&  20.16& 79.23 & 44.95& 0.227 \\
& \checkmark & $\times$ & 21.49& 80.73 & 49.97& 0.175 \\
 &$\times$ & \checkmark & \underline{24.58}& \underline{82.71} & \underline{51.44}& \underline{0.164} \\
 & \checkmark & \checkmark  & \textbf{28.20} &\textbf{83.64} & \textbf{58.80} & \textbf{0.157} \\ 
 \midrule
 \multirow{4}{*}{\rotatebox{90}{Unseen}} 
  &$\times$ & $\times$  &  7.120& 66.35 & 32.11& 0.301 \\
& \checkmark & $\times$ & 8.010& 69.13 & 33.26& 0.295 \\
 &$\times$ & \checkmark & \underline{9.520}& \underline{70.11} & \underline{36.46}& \underline{0.273} \\
 & \checkmark & \checkmark  & \textbf{10.95} &\textbf{72.86} & \textbf{40.08} & \textbf{0.255} \\ 
 
 \bottomrule
\end{tabular}}

\label{ablation1}
\end{table}

\subsection{Benchmark Setting}
\label{benchmark_setting}
\paragraph{Baselines.} As this is a newly introduced task that targets extracting affordance knowledge within HOI videos and transferring it for 3D affordance grounding, there are no previous works. The most related work to ours is EGO-SAG, while the code has not been released. For a thorough comparison of our method, we select several advanced HOI images based on 3D affordance grounding methods (e.g., IAGNet~\cite{IAGNet}, GREAT~\cite{GREAT}, AGILE~\cite{AGPIL}) as modular baselines. We apply the same frames sampling strategy to get N frames, and we then utilize the image encoder used in these methods to encode each frame and finally fuse the embeddings. 

\paragraph{Evaluation Metrics.}Following previous works, we chose four evaluation metrics: Area Under the Curve (AUC)~\citep{lobo2008auc}, Mean Intersection Over Union (mIOU)~\citep{rahman2016optimizing}, SIMilarity (SIM)~\citep{swain1991color}, and Mean Absolute Error (MAE)~\citep{willmott2005advantages}.

\subsection{Implementation Details.}
\label{implen}
Following VideoLLaVA~\cite{lin2023video}, we utilize LanguageBind~\cite{zhu2023languagebind} as the video encoder, Llama~\cite{touvron2023llama} as the large language model. For the action encoder, we employ RenderNet~\cite{liu2025stamo} as the latent action encoder due to its powerful world modeling capability, and we adopt Uni3D~\cite{zhang2023uni3d} as the 3D vision encoder to enhance the 3D understanding ability. All the encoders are frozen.  We employ LoRA~\cite{hu2021lora} for efficient fine-tuning and set the rank of LoRA to 128 by default. Additionally, we utilize AdamW~\cite{loshchilov2017decoupled} optimizer with the learning rate and weight decay set to 0.0002 and 0, respectively. We adopt a cosine learning rate scheduler, with the warm-up iteration ratio set to 0.03. All attentions in our model are replaced by flash-attention~\cite{dao2022flashattention} during training. The training is done on four H200 GPUs for 10 epochs for the main experiments, and during training, the overall training process takes nearly 11 hours. More details can be seen in the Appendix.
\label{sec:exe}

\begin{figure*}[t]
    \centering
    \includegraphics[width=\linewidth]{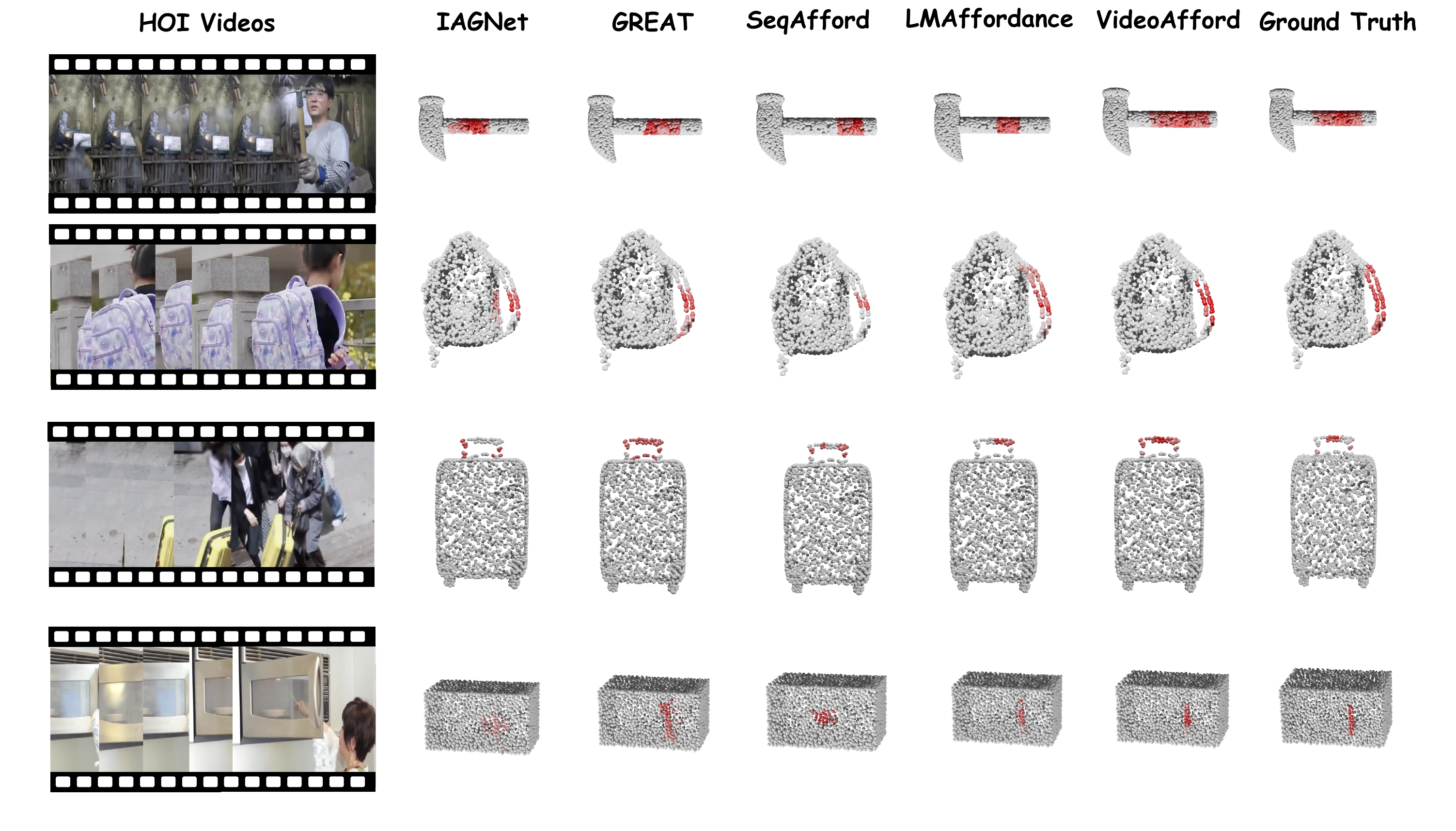}
    \caption{\textbf{Visualization Results.} The first column is the HOI videos, and the last column is the ground truth of 3D object affordance in the point
cloud. The depth of red represents the affordance probability. Refer to our supplementary materials for more results.}
    \label{fig:vis}
\end{figure*}

\begin{table}[t]
\centering
\caption{We conduct the experiments to figure out the influence of different sampled video frames. The best results are in \textbf{bold} and the second results are in \underline{underline}.}
\resizebox{\linewidth}{!}{

\begin{tabular}{@{}lccccc@{}}  
\toprule
& \textbf{Metric} & \textbf{2 frames} & \textbf{4 frames} & \textbf{8 frames} & \textbf{16 frames} \\
\midrule
 \multirow{4}{*}{\rotatebox{90}{Seen}} 
& \textit{mIoU}$\uparrow$   & 22.30 & \underline{24.22} & \textbf{28.20} & 18.53\\  
& \textit{AUC}$\uparrow$    & 80.12 & \underline{82.61} & \textbf{83.64} & 77.48\\
& \textit{SIM}$\uparrow$    & 50.74 & 51.96 & \textbf{58.80} & 41.93\\
& \textit{MAE}$\downarrow$  & \underline{0.183}  & 0.185  & \textbf{0.157} & 0.197\\
\midrule
\multirow{4}{*}{\rotatebox{90}{Unseen}} 

& \textit{mIoU}$\uparrow$   & 8.510 & \underline{9.770} & \textbf{10.95} & 6.720\\  
& \textit{AUC}$\uparrow$    & 69.50 & \underline{71.19} & \textbf{72.86} & 69.24\\
& \textit{SIM}$\uparrow$    & 36.17 & \underline{39.82} & \textbf{40.08} & 37.12\\
& \textit{MAE}$\downarrow$  & 0.260  & \underline{0.258}  & \textbf{0.255} & 0.275\\ 

\midrule
\bottomrule
\end{tabular}}

\label{tab:abla_frames}
\end{table}
\subsection{Comparisoni Results}
\label{Comparison Results}
 Table~\ref {main_results} shows the results of different affordance grounding approaches on the proposed dataset VIDA, which demonstrate that our method consistently outperforms all other approaches across all evaluated metrics. The baseline model, which produces poor results, suggests that simply encoding the frames of the video is insufficient to capture the dynamic interaction and intricate correlations between videos and 3D sources. Thanks to the powerful video understanding and latent action modeling capabilities of VideoAfford, our model is able to capture dynamic information in HOI videos, showing strong generalization, and extract the general affordance knowledge within HOI videos. In the Seen setting, all the objects and affordance types are ``seen" by our model, and in the Unseen setting, the training set does not include some objects, which causes a huge challenge for the grounding models to predict the affordance area.  All the baselines fail to predict good results, while our method anticipates precise results by mining affordance clues provided by dynamic interactions. Additionally, the visual qualitative comparative results of our method and other baselines
are shown in Fig.\ref{fig:vis}. As can be seen, the comparative baselines could anticipate some 3D object affordance under our setting, but in comparison, our method obviously achieves better results, which validates the rationality of our setting. When provided with an HOI video, our model can understand how the HOI video is connected with actionable affordances and accurately extract the dynamic interaction knowledge from the video. This ability is not only attributed to the challenging benchmark collected from diverse sources but also to the powerful world knowledge internalized in video MLLMs.

\subsection{Ablation Study}
\label{abla}

In this section, we conduct a comprehensive ablation study to investigate the effect of different framework designs, loss function design, and hyperparameter settings. 

\paragraph{ Effectiveness of Action Encoder and Spatial Loss.} As shown in Table~\ref{ablation1}, it reports the impact of the action encoder and spatial loss. Introducing these modules results in a substantial improvement over the baseline, which underscores that our method enables deep integration of rich dynamic interaction priors within HOI videos, rich affordance world knowledge within the video LLM. Without the action encoder, the model fails to capture the latent actions, resulting in incorrect predictions in irrelevant areas and a marked decline in both performance and efficiency. Without Spatial Loss constraints, the model only focuses solely on ``whether the predicted class for each point is accurate," but lacks constraints on ``spatial continuity and regional overlap.", resulting in a noticeable drop in the IOU metrics.

\paragraph{ Choice of Sampled Frames.} We conducted an ablation study to systematically investigate the impact of sampled frames on model performance. As presented in Table \ref{tab:abla_frames}, when only 2 or 4 frames are sampled, the model struggles to capture the complete interaction dynamics, leading to suboptimal results across key metrics. Conversely, an excessive number of sampled frames (e.g., 16 frames) not only introduces redundant, cluttered information that interferes with effective feature learning but also incurs higher computational overhead. Sampling 8 frames achieves a balance between capturing sufficient temporal context and avoiding information overload; thus, we adopt this as the default configuration for both seen and unseen experiments.


\section{Conclusion and Future Works}
In this paper, we introduce the task \textit{Grounding 3D object affordance from human-object interaction demonstration videos}, which aims to harness large-scale demonstration video corpora to advance object-centric 3D affordance reasoning from static interaction knowledge to complex, dynamic interaction priors. We collect and construct the first large-scale video-based 3D object affordance dataset \textbf{VIDA}, and based on this strong dataset, we introduce \textbf{VideoAfford}, the first MLLM to reason fine-grained 3D affordance for this new paradigm. Bolstered by novel Spatial Constraint Loss and the powerful world knowledge within MLLM, our method establishes a strong baseline for this new but meaningful task, and we will continue to explore how to unleash the huge potential of unlabeled HOI videos for embodied perception.

\section*{Impact Statement}
This paper presents work whose goal is to advance the field of Machine Learning. There are many potential societal consequences of our work, none of which we feel must be specifically highlighted here.

\nocite{langley00}

\bibliography{example_paper}
\bibliographystyle{icml2026}

\end{document}